\documentclass[10pt,twocolumn,letterpaper]{article}

\usepackage[pagenumbers]{cvpr} %

\usepackage[dvipsnames]{xcolor}

\usepackage{amssymb}
\usepackage{xcolor}
\usepackage{comment}
\usepackage{subcaption}
\usepackage{tabularx}
\usepackage{stmaryrd}
\usepackage{mathrsfs}
\usepackage{url}
\usepackage{pifont}
\usepackage{algorithm}
\usepackage{algpseudocode}
\usepackage{braket}
\usepackage{stfloats}

\newcommand{\cmark}{\ding{51}}%
\newcommand{\xmark}{\ding{55}}%
 
\definecolor{ColorGray}{RGB}{96, 96, 96}

\newcommand{\myparagraph}[1]{\vspace{1mm}\noindent\textbf{#1}~}
\newcommand{\myitem}[1]{\vspace{1mm}\noindent#1~}

\def\OurName{GenZI}

\def\InScene{\mathcal{S}}
\def\InHuman{\mathcal{B}}
\def\InText{\varGamma}
\def\InLocation{\mathbf{p}}

\def\InHumanVertices{\mathbf{V}}
\def\InHumanVertex{\mathbf{v}}
\def\InHumanFaces{\mathbf{F}}

\def\CamDistance{d}
\def\PatchRadius{r}

\def\VarGlobalTranslation{\mathbf{t}}
\def\VarGlobalRotation{\mathbf{R}}
\def\VarBodyPose{\mathbf{\Theta}}
\def\VarBodyShape{\mathbf{\Phi}}
\def\VarViewsConsistency{\mathbf{w}}
\def\VarViewConsistency{w}

\def\JointAngles{\hat{\mathbf{\Theta}}}

\def\SceneImage{\mathbf{I}}
\def\SceneHumanImage{\bar{\mathbf{I}}}
\def\NumSceneImages{k}
\def\SDInpaint{\varOmega}
\def\SDNoisyImage{\mathbf{z}}
\def\SDMask{\mathbf{M}}
\def\SDAttentionMaps{\mathbf{A}}
\def\SDNumPixels{hw}
\def\SDNumTokens{n}
\def\SDHumanTokens{h}

\def\TwoDJoints{\mathbf{J}}
\def\TwoDJointsConfidence{\mathbf{c}}
\def\ThreeDJoints{\hat{\mathbf{J}}}

\def\CameraProjection{\varPi}
\def\ViewConsistencyNum{\tau}
\def\SceneSDF{\varPsi}

\def\JointFitEnergy{\mathcal{E}_{\text{PF}}}
\def\JointFitWeight{\lambda_{\text{PF}}}

\def\ViewConsistencyEnergy{\mathcal{E}_{\text{VS}}}
\def\ViewConsistencyWeight{\lambda_{\text{VS}}}

\def\BodyPoseEnergy{\mathcal{E}_{\text{BP}}}
\def\BodyPoseWeight{\lambda_{\text{BP}}}

\def\JointAngleEnergy{\mathcal{E}_{\text{JA}}}

\def\BodyShapeEnergy{\mathcal{E}_{\text{BS}}}
\def\BodyShapeWeight{\lambda_{\text{BS}}}

\def\PhysicalEnergy{\mathcal{E}_{\text{SC}}}
\def\PhysicalWeight{\lambda_{\text{SC}}}

\def\SelfPenetrationEnergy{\mathcal{E}_{\text{SP}}}
\def\SelfPenetrationWeight{\lambda_{\text{SP}}}

\def\TotalEnergy{\mathcal{E}_{\text{total}}}

\definecolor{cvprblue}{rgb}{0.21,0.49,0.74}
\usepackage[pagebackref,breaklinks,colorlinks,citecolor=cvprblue]{hyperref}

\def\paperID{***} %
\def\confName{***}
\def\confYear{***}

\title{\OurName{}: Zero-Shot 3D Human-Scene Interaction Generation}

\author{Lei Li \hspace{1cm} Angela Dai\\
Technical University of Munich\vspace{0.1cm}\\
\href{https://craigleili.github.io/projects/genzi/}{craigleili.github.io/projects/genzi}
\vspace{-0.5cm}
}

\begin{document}

\twocolumn[{%
    \maketitle
    \begin{center}
        \centerline{\includegraphics[width=\linewidth]{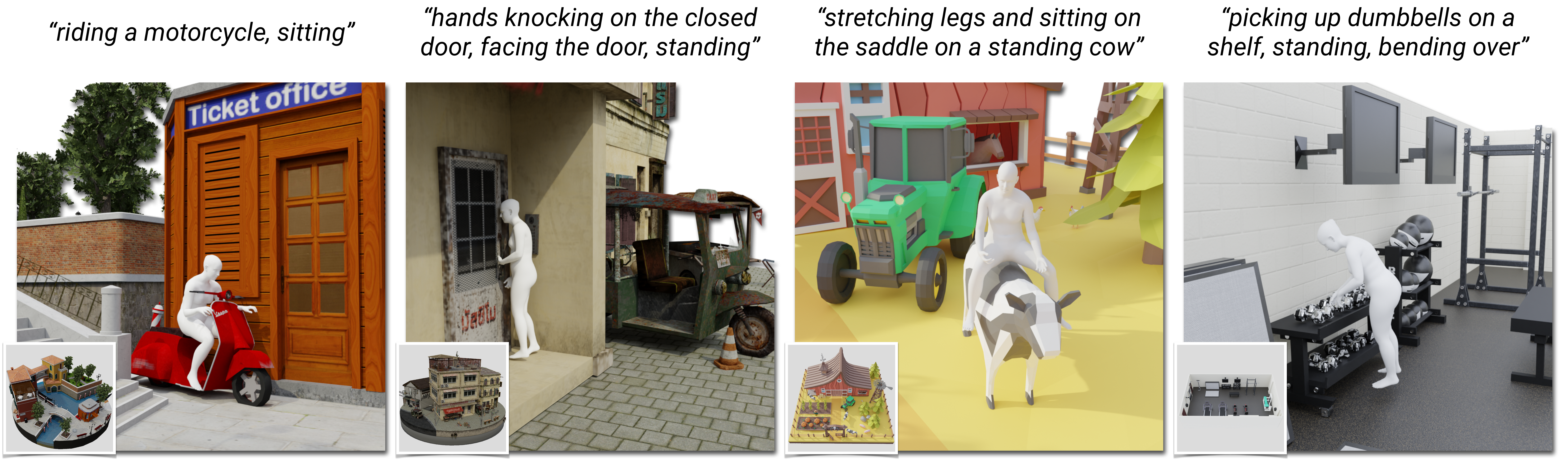}}
        \vspace{-0.2cm}
        \captionof{figure}{Given an arbitrary 3D scene, \OurName{} can synthesize virtual humans interacting with the 3D environment at specified locations from a brief text description. Our approach does \emph{not} require any 3D human-scene interaction training data or 3D learning. By distilling interaction priors from powerful 2D vision-language models, we optimize for 3D human-scene interaction synthesis in a flexible fashion, with simple language-based control and high generality to various types of scene environments.}
        \label{fig_teaser}
    \end{center}%
}]

\begin{abstract}
    \vspace{-0.1cm}
    Can we synthesize 3D humans interacting with scenes without learning from any 3D human-scene interaction data? We propose \OurName{}, the first zero-shot approach to generating 3D human-scene interactions. Key to \OurName{} is our distillation of interaction priors from large vision-language models (VLMs), which have learned a rich semantic space of 2D human-scene compositions. Given a natural language description and a coarse point location of the desired interaction in a 3D scene, we first leverage VLMs to imagine plausible 2D human interactions inpainted into multiple rendered views of the scene. We then formulate a robust iterative optimization to synthesize the pose and shape of a 3D human model in the scene, guided by consistency with the 2D interaction hypotheses. In contrast to existing learning-based approaches, \OurName{} circumvents the conventional need for captured 3D interaction data, and allows for flexible control of the 3D interaction synthesis with easy-to-use text prompts. Extensive experiments show that our zero-shot approach has high flexibility and generality, making it applicable to diverse scene types, including both indoor and outdoor environments.
\end{abstract}

\section{Introduction}
\label{sec_introduction}

Scenes are typically constructed to enable human interactions with the environment, like sitting on a couch, playing a piano, or opening a mailbox. Understanding these interactions between humans and scenes, also known as affordances \cite{koffka1935principles,gibson1978ecological}, has gained increasing attention in the fields of computer vision and graphics. In particular, achieving controllable and generalizable synthesis of human-scene interactions (or HSIs) holds immense potential for various applications, such as robotics, architectural design, video games, and virtual reality experiences, among many others.

Generating realistic humans interacting with a 3D scene is a challenging task, requiring holistic semantic understanding of both the environment and possible human actions therein. Existing approaches to HSI synthesis \cite{Zhang_2020_CVPR,Hassan_2021_CVPR,zhao2022compositional,xuan2023narrator} rely heavily on supervised training using meticulously  captured data of real people interacting in 3D environments. Unfortunately, collecting large-scale datasets of 3D scenes and human interactions is exorbitantly difficult. It not only demands accurate tracking and reconstruction of both people and their environments, but also needs to ensure sufficient diversity in subjects and scenes. Existing HSI datasets currently contain very limited quantities of scenes and actions, for example, PROX \cite{Hassan_2019_ICCV,zhao2022compositional} only consisting of 12 indoor scenes and humans interacting with 11 object categories.
The scarcity of ground truth data for supervision has thus strongly limited the applicability and generalization of the learning-based approaches for synthesizing diverse sets of actions in arbitrary 3D scenes.

We thus consider an alternative perspective to HSI synthesis and pose the following question: \emph{Can we achieve plausible HSI synthesis without using any captured 3D interaction data?} To this end, we present a novel \emph{zero-shot} approach to 3D HSI generation. We propose the first method to leverage the powerful capabilities exhibited by recent vision-language models (VLMs) \cite{rombach2022high,saharia2022photorealistic,radford2021clip,li2022blip,li2023blip2} to synthesize plausible 2D images of human interactions, and introduce a robust optimization to distill inferred 2D pose information into 3D human synthesis in a 3D scene.

More concretely, given a 3D scene, a text prompt and a coarse point location of the desired interaction, \OurName{} optimizes for the pose and shape of a 3D human performing the action in the scene, guided by a large VLM \cite{rombach2022high}. We first leverage the VLM to imagine possible 2D humans by inpainting images from multiple rendered views of the scene. We automate this 2D human insertion process with a dynamic masking scheme that automatically updates proposed masks through the inpainting process, eliminating the need for manual specifications of human inpainting regions. We then lift these 2D interaction hypotheses to 3D and optimize for a parametric 3D human body model \cite{loper2015smpl, pavlakos2019expressive} that is most consistent with the 2D pose guidance. We further refine the generated 3D human in the scene by iterating through the VLM-based 2D inpainting and robust 3D lifting stages. We demonstrate the flexibility and generality of \OurName{} in various types of 3D environments (\cref{fig_teaser}), encompassing both indoor and outdoor scenes.

In summary, our contributions are as follows:
\begin{itemize}
\item We introduce \OurName{}\footnote{Our code and data will be made publicly available upon publication.}, the first zero-shot approach to generate realistic 3D humans interacting with a 3D scene from natural language prompts. \OurName{} does not require any supervision from 3D interaction data, thus enabling flexible synthesis across diverse scenes and actions.
\item We propose a dynamically-masked inpainting scheme that allows for the synthesis of plausible 2D human-scene compositions via VLMs without requiring manually-specified human inpainting masks.
\item We develop a robust 3D pose optimization to lift various inferred images of human interactions to a view-consistent, realistic 3D HSI synthesis.
\end{itemize}

\section{Related Work}
\label{sec_related_work}

\begin{figure*}[t]
    \centering
    \includegraphics[width=\textwidth]{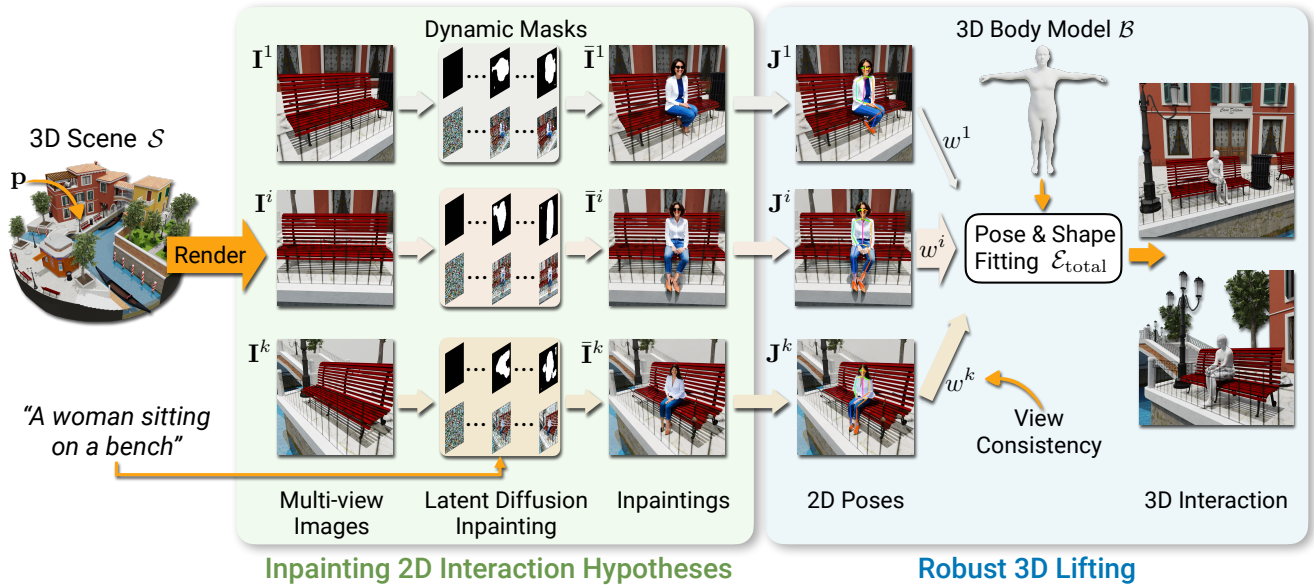}
    \caption{
    \OurName{} distills information from vision-language model for 3D human-scene interaction.
    We first leverage large vision-language models to synthesize possible 2D humans interactions with the 3D scene $\InScene$ by employing latent diffusion inpainting \cite{rombach2022high} on multiple rendered views of the environment at location $\InLocation$ using our dynamic masking scheme to automatically estimate inpainting masks.
    We then lift these 2D hypotheses to 3D in a robust optimization for a 3D parametric body model $\InHuman$ (SMPL-X~\cite{pavlakos2019expressive}) that is most consistent with detected 2D poses in the inpainted 2D hypotheses.
    This produces a semantically consistent interaction that respects the scene context, without requiring any 3D human-scene interaction data.}
    \label{fig_pipeline}
\end{figure*}

\myparagraph{3D Human-Scene Interaction Synthesis.}
Synthesizing humans in scenes is an important, challenging task in computer vision and graphics, as it models complex high-level semantic understanding such as affordances and interactions.
Existing approaches for human-scene interaction synthesis \cite{savva2016pigraphs,monszpart2019imapper,Zhang_2020_CVPR,Hassan_2021_CVPR,zhao2022compositional} focus on learning prior knowledge from available scanned data of people interacting with 3D indoor scenes \cite{Hassan_2019_ICCV} in a supervised manner. 

The early work PiGraphs \cite{savva2016pigraphs} introduced a probabilistic graphical model of human-object interactions, generating static human-object placements by pose sampling and model retrieval. Zhang \etal \cite{Zhang_2020_CVPR} introduced a generative model of human-scene interaction, using a conditional variational autoencoder (CVAE) to model the distribution of 3D human poses conditioned on scene depth and semantics. PLACE~\cite{zhang2020place} explicitly represented human-scene interactions with Basis Point Sets (BPS) encoding \cite{prokudin2019efficient}, training a CVAE to synthesize such representations for generating natural human poses and contact relations within scenes. POSA~\cite{Hassan_2021_CVPR} proposed an ego-centric representation by augmenting the SMPL-X model \cite{pavlakos2019expressive} with contact labels and scene semantics, using  a CVAE conditioned on SMPL-X vertex positions to model potential interactions. Recently, COINS~\cite{zhao2022compositional} presented a method for compositional human-scene interaction synthesis with high-level semantic control, using transformer-based CVAEs conditioned on provided 3D objects and interaction semantics to regress human body poses and contact features.

However, these works all require supervision from 3D human-scene interactions -- such datasets are difficult and expensive to capture and annotate, for instance, the widely used PROX dataset \cite{Hassan_2019_ICCV,zhao2022compositional} only has 8 reconstructed indoor scenes for training and 4 for testing, capturing human interactions with 11 object categories.
As such, these methods are limited to in-domain synthesis, prohibiting applicability to general 3D scene settings with arbitary objects and arrangements. 
In this work, we deviate from the learning-based approach and propose to take advantage of established large vision-language models \cite{rombach2022high,saharia2022photorealistic,radford2021clip,li2023blip2} for human-scene interaction synthesis, thereby bypassing the requirement for data capture and 3D learning.

\myparagraph{3D Human Estimation from RGB Images.}
Over the past decades, significant progress has been made in 3D human estimation from RGB images \cite{gavrila1999visual,moeslund2006survey,poppe2007vision,sarafianos20163d}. Among those, the prominent work SMPLify-X~\cite{pavlakos2019expressive} fits the SMPL-X model to 2D joints estimated from a single image through optimization. Hassan \etal~\cite{Hassan_2019_ICCV} build upon SMPLify-X and estimate 3D human poses by incorporating additional physical contact and penetration contraints between human and 3D scene. Learning-based approaches \cite{kolotouros2019learning,choutas2020monocular,rong2021frankmocap,zhou2021monocular,zhang2023pymaf,lin2023one,cai2023smpler} have received much research attention in recent years towards tackling 3D human estimation from monocular images with improved hand and face estimation. 
Our work leverages 2D pose reasoning from multiple different 2D view hypotheses, and proposes a robust 3D formulation for aggregating the various 2D hypotheses to a consistent 3D human body interacting with a 3D scene.

\myparagraph{Distilling Prior from Vision-Language Models.}
Recent advances in powerful vision-language models \cite{rombach2022high,saharia2022photorealistic,radford2021clip,li2023blip2} have also inspired various works aiming to distill information learned through the models to various tasks, including 2D panoptic segmentation \cite{xu2023odise}, 3D semantic segmentation \cite{rozenberszki2022language}, 3D scene generation \cite{hoellein2023text2room}, and synthesis of images with hand-object interactions \cite{ye2023affordance}.
Our approach leverages the 2D generative capacity of these models to convey information about possible human-scene interactions, which we then lift to a 3D, consistent interaction.

\section{Method}
\label{sec_method}

\subsection{Overview}
Our objective is to synthesize plausible 3D humans interacting with a 3D scene, guided by input text descriptions, in the absence of 3D interaction data capture for learning. We present \OurName{}, a novel optimization-based multi-view approach that leverages large VLMs to infer spatial relations of interactions between a human and the scene. \cref{fig_pipeline} shows an illustration of our approach.

\OurName{} takes as input a 3D scene $\InScene$, a text prompt $\InText$ describing the desired interaction, and an approximate point location $\InLocation\in\mathbb{R}^3$ in the scene around which the interaction should occur. Our approach generates a posed 3D human $\InHuman$ as output, performing the specified action in the scene. We adopt SMPL-X \cite{pavlakos2019expressive,loper2015smpl} to parameterize the 3D human $\InHuman$, as it provides a fully differentiable function mapping a set of pose and shape parameters $(\VarGlobalRotation, \VarGlobalTranslation, \VarBodyPose, \VarBodyShape)$ to a 3D human mesh with vertices $\InHumanVertices$ and faces $\InHumanFaces$.
We thus optimize for $(\VarGlobalRotation, \VarGlobalTranslation, \VarBodyPose, \VarBodyShape)$ characterizing $\InHuman$ in the scene $\InScene$, with $\VarGlobalRotation \in \mathbb{R}^{6}$ denoting the global orientation as a continuous rotation representation \cite{zhou2019continuity}, $\VarGlobalTranslation \in \mathbb{R}^{3}$ the global translation, $\VarBodyPose \in \mathbb{R}^{32}$ the body pose in the latent space of VPoser \cite{pavlakos2019expressive} which decodes to body joint rotations $\JointAngles \in \mathbb{R}^{21 \times 3}$, and $\VarBodyShape\in \mathbb{R}^{10}$ representing the body shape as blend shape coefficients.

\OurName{} synthesizes the desired interaction between the 3D human $\InHuman$ and the scene $\InScene$ by distilling information from VLMs through 2D human inpainting followed by robust 3D lifting. We first generate a collection of plausible 2D human-scene compositions by employing a large VLM to inpaint humans into multiple rendered views of the scene $\InScene$ (\cref{subsec_multiview_interaction_inpainting}). We then introduce a robust 3D lifting procedure that optimizes for the pose and shape of the human $\InHuman$, guided by consistency with the 2D interaction hypotheses (\cref{subsec_3d_interaction_optimization}). We further refine the posed 3D human $\InHuman$ by iteratively updating the 2D inpaintings and 3D optimization.

\subsection{Inpainting Multi-view Interaction Hypotheses}
\label{subsec_multiview_interaction_inpainting}

We first leverage a VLM to generate 2D hypotheses of potential human interactions in the scene by automatically inpainting humans into multiple rendered views of $\InScene$.

\myparagraph{Multi-view Rendering.} 
To capture scene context for 2D human inpainting, we render multiple views of the 3D scene $\InScene$ from $\NumSceneImages$ virtual cameras looking at $\InLocation$. 
Cameras are randomly sampled on a hemisphere and filtered according to the visibility of $\InLocation$; we refer to the supplementary material for additional camera setup details.
We denote the rendered scene images as $\{ \SceneImage^{i} \}_{i \in [1, \NumSceneImages]}$. To simplify notation, we omit the superscript $i$ indexing each view in this section.

\myparagraph{Inpainting with Dynamic Masking.} Given a rendered scene image $\SceneImage$ and the text prompt $\InText$, we leverage a state-of-the-art latent 2D diffusion model \cite{rombach2022high} to generate a new image $\SceneHumanImage$, where a human is inpainted into the scene image $\SceneImage$ while adhering to the specified interaction and 2D scene context. In practice, we opt for the popular Stable Diffusion Inpainting model \cite{stablediffusioninpainting} as the latent diffusion implementation. 

The latent diffusion model, denoted as $\SDInpaint(\SDNoisyImage_{t}, \SDMask, \SceneImage, \InText, t)$, performs image inpainting by progressively denoising a noisy latent $\SDNoisyImage_{t}$ at each time step $t$. During the denoising diffusion process, the binary mask $\SDMask$ defines the inpainting region in the image $\SceneImage$; however, specifying this mask typically requires manual effort \cite{stablediffusioninpainting}. Thus, we develop a fully-automated inpainting process by automatically generating the mask. 
Note that using a random or fixed human mask naively can lead to incorrect inpaintings, as this often results in scene context incorrectly masked, \eg, the objects affording the desired action may be entirely masked out, leading to the generated 2D human poses to be incoherent with the scene, producing undesirable 3D HSI synthesis.

We propose a masking scheme that dynamically adapts the mask through the denoising process by leveraging the internal cross-attention maps \cite{vaswani2017attention,hertz2023prompttoprompt,chefer2023attend} from $\SDInpaint$ to propose masks.
The cross-attention maps capture rich semantic correlations between image pixels and input text tokens, playing a crucial role in guiding image generation. Let $\SDAttentionMaps \in \mathbb{R}^{\SDNumPixels \times \SDNumTokens}$ denote the cross-attention map between an image feature map of $\SDNumPixels$ pixels and the text prompt $\InText$ with $\SDNumTokens$ tokens, normalized row-wise using {\tt softmax}. Here, $\SDAttentionMaps[i, j]$ signifies the influence of the $j$-th token on the $i$-th pixel, and $\SDAttentionMaps[:, j]$ forms a heat map of image regions to be filled with content related to the $j$-th token. 

\begin{figure}[t]
    \centering
    \includegraphics[width=\columnwidth]{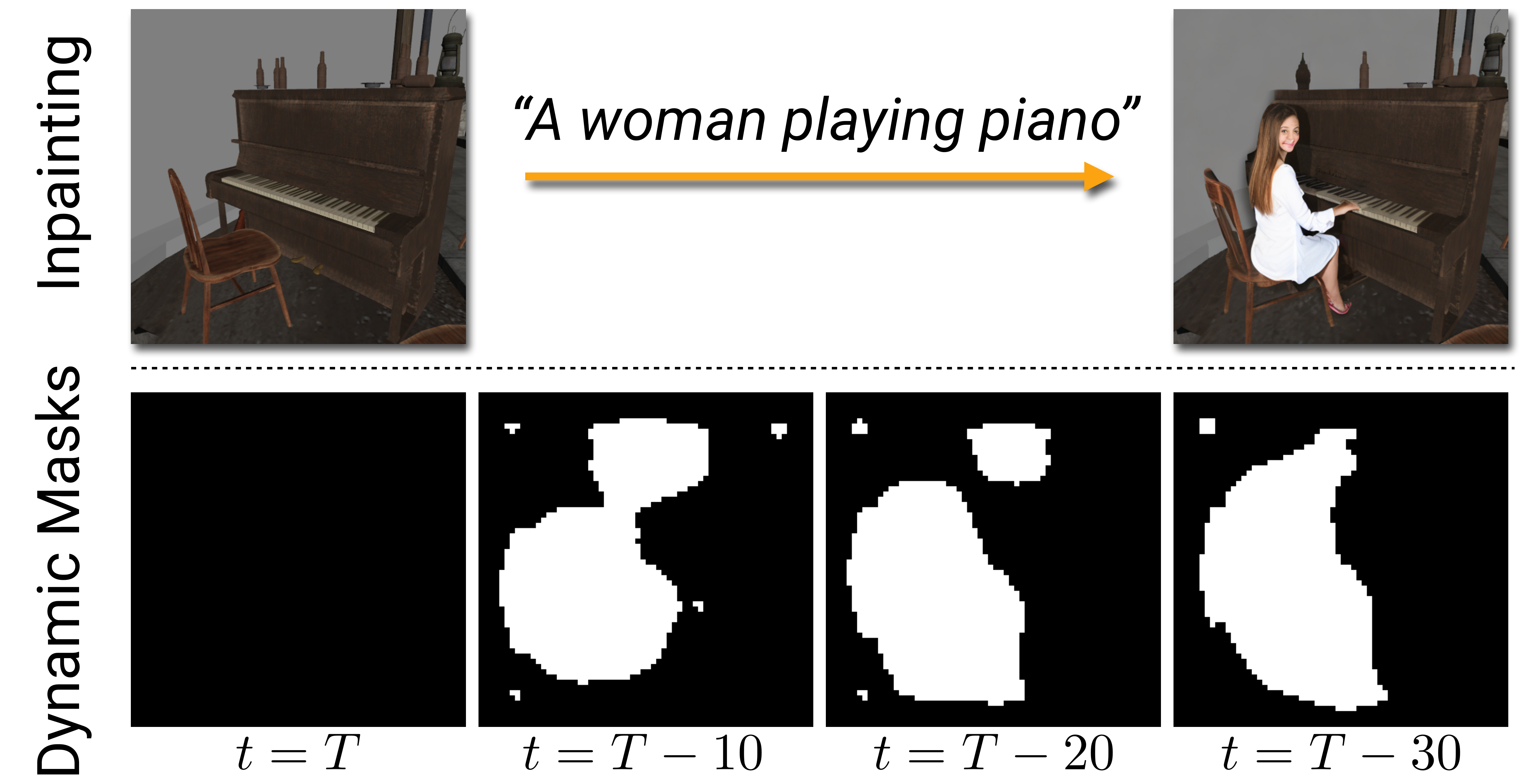}
    \caption{Human inpainting with dynamic masking. Top: Given a scene image and a text prompt, a human is inpainted into the image \emph{without} a mask specifying the inpainting region for latent diffusion. Bottom: The masks generated by our dynamic masking scheme based on cross-attention maps at different diffusion time steps adaptively shift to find the region of interest.}
    \label{fig_dynamic_masking}
\end{figure}

Using the cross-attention map $\SDAttentionMaps_{t}$ at each time step $t$ of the diffusion process, we can dynamically derive a mask related to human tokens. Specifically, at time step $t$, we obtain $\SDNoisyImage_{t-1}, \SDAttentionMaps_{t} \leftarrow \SDInpaint(\SDNoisyImage_{t}, \SDMask_{t}, \SceneImage, \InText, t)$ after denoising. To create a human inpainting mask $\SDMask_{t-1}$ for time step $t-1$, we extract heat maps from $\SDAttentionMaps_{t}$ corresponding to the tokens referring to the human (\eg, ``woman'' or ``man''), followed by summation of these extracted heat maps across the tokens and binarization. At the initial time step $T$, we initialize $\SDMask_{T}$ as an empty mask. Our dynamic masking approach enables the synthesis of a 2D human-scene composition image $\SceneHumanImage$ without the need for manually-specified human inpainting masks. We illustrate our dynamic masking scheme in \cref{fig_dynamic_masking}.

\subsection{Robust Lifting to a 3D Interaction}
\label{subsec_3d_interaction_optimization}

Given the multi-view scene images inpainted with human subjects $\{ \SceneHumanImage^{i} \}_{i \in [1, \NumSceneImages]}$, our aim is to optimize for the pose and shape parameters $(\VarGlobalRotation, \VarGlobalTranslation, \VarBodyPose, \VarBodyShape)$ of a 3D human $\InHuman$, guided by the multi-view interaction cues. We optimize for $\InHuman$ by matching it with the 2D poses extracted from $\{ \SceneHumanImage^{i} \}$. Since the 2D hypotheses may not be consistent across views, we formulate this as a robust optimization, simultaneously optimizing for the set of views most consistent with $\InHuman$.

\myparagraph{2D Pose Estimation}
To distill interaction guidance for the 3D HSI synthesis, we compute a 2D pose representation for the inpainted human in each image $\SceneHumanImage^{i}$. We use AlphaPose \cite{Fang2022alphapose}, an off-the-shelf pose estimation approach that infers a set of 2D joint positions $\TwoDJoints^{i}$ and the corresponding joint confidence scores $\TwoDJointsConfidence^{i}$ for the human subject in image $\SceneHumanImage^{i}$. 

These 2D pose hypotheses $\{ \TwoDJoints^{i}, \TwoDJointsConfidence^{i} \}_{i \in [1, \NumSceneImages]}$ are then used to steer the interaction synthesis between the 3D human $\InHuman$ and the scene $\InScene$. We aim to minimize the following objective function $\TotalEnergy$:
\begin{equation}
    \label{eq_total_energy}
    \begin{split}
        \TotalEnergy = &\JointFitWeight \JointFitEnergy + \ViewConsistencyWeight \ViewConsistencyEnergy + \BodyPoseWeight \BodyPoseEnergy + \\
        &\BodyShapeWeight \BodyShapeEnergy + \PhysicalWeight \PhysicalEnergy + \SelfPenetrationWeight \SelfPenetrationEnergy,
    \end{split}
\end{equation}
where the $\lambda$ denote scalar weights for the energy terms: pose fitting $\JointFitEnergy$, view selection $\ViewConsistencyEnergy$, body pose $\BodyPoseEnergy$, body shape $\BodyShapeEnergy$, scene penetration $\PhysicalEnergy$, and self-penetration $\SelfPenetrationEnergy$.

\myparagraph{Robust View-consistent 3D Pose Fitting.} Our primary energy term $\JointFitEnergy$ minimizes the discrepancy between the projections of the 3D pose of $\InHuman$ and the inpainted poses from multiple views. 
However, 2D pose hypotheses can often be inconsistent across different views, due to the stochastic nature of diffusion models, leading to conflicting optimization signals for $\InHuman$. 
To address this, we employ a robust optimization strategy that optimizes additionally for view selection weights in $\JointFitEnergy$ that promote consistency between the 3D pose and the most consistent 2D pose hypotheses. 

We thus introduce a new set of \emph{optimizable} variables $\VarViewsConsistency = \{ \VarViewConsistency^{i} | \VarViewConsistency^{i} \in [0, 1] \}_{i \in [1, \NumSceneImages]}$, representing view consistency scores, and apply a robust kernel $\rho$ to the per-view joint fitting constraints. The weights $\VarViewsConsistency$ allow the solver to adaptively focus on views with consistent inpainted poses and downweight the inconsistent ones. Our pose fitting energy $\JointFitEnergy$ with view consistency is formulated as follows:
\begin{equation}
    \label{eq_joint_fit_energy}
    \JointFitEnergy = \frac{\sum_{i} \VarViewConsistency^{i} \sum_{j} \TwoDJointsConfidence^{i}_{j} \rho \big(\CameraProjection(\ThreeDJoints)_{j}^{i} - \TwoDJoints_{j}^{i} \big)}{\sum_{i} \VarViewConsistency^{i}},
\end{equation}
where $\ThreeDJoints$ denotes the 3D joint positions of the SMPL-X model \cite{pavlakos2019expressive} differentiably computed from the pose and shape parameters $(\VarGlobalRotation, \VarGlobalTranslation, \VarBodyPose, \VarBodyShape)$. The function $\CameraProjection(\cdot)_{j}^{i}$ represents projection of the $j$-th joint in the $i$-th camera view, and $\rho$ is the robust Geman-McClure error function \cite{Geman1987StatisticalMF}.

\myparagraph{Regularizations.} Below, we describe the regularization terms adopted in our optimization.

\myitem{1)} We encourage the per-view weights $\VarViewsConsistency$ to focus on at least $\ViewConsistencyNum$ views:  
\begin{equation}
    \label{eq_view_consistency_energy}
    \ViewConsistencyEnergy = \max(\ViewConsistencyNum - \sum_{i} \VarViewConsistency^{i}, 0).
\end{equation}

\myitem{2)} To ensure a natural 3D pose, we impose the following energy term on the body pose parameter:
\begin{equation}
    \label{eq_body_pose_energy}
    \BodyPoseEnergy = \| \VarBodyPose \|^{2} + \JointAngleEnergy(\JointAngles).
\end{equation}
The first term is the VPoser body prior \cite{pavlakos2019expressive} regularizing the latent pose $\VarBodyPose$, and the second term $\JointAngleEnergy$ is a simple angle prior on the body joint rotations $\JointAngles$, decoded from the latent pose $\VarBodyPose$ by VPoser, to penalize extreme bending. The formulation of $\JointAngleEnergy$ is provided in the supplementary material.

\myitem{3)} We reguarlize the shape parameter $\VarBodyShape$ to obtain a plausible body shape via
\begin{equation}
    \label{eq_body_shape_energy}
    \BodyShapeEnergy = \| \VarBodyShape \|^{2},
\end{equation}
which measures the Mahalanobis distance between $\VarBodyShape$ and the body shape distribution used in SMPL-X \cite{pavlakos2019expressive}.

\myitem{4)} To ensure physical contact but also avoid penetration between the 3D human $\InHuman$ and the scene $\InScene$, we formulate the spatial constraints as:
\begin{equation}
    \label{eq_physical_energy}
    \PhysicalEnergy = 
    \begin{cases}
        \min_{\InHumanVertex \in \InHumanVertices} \SceneSDF(\InHumanVertex),  &  \SceneSDF(\InHumanVertex) > 0 \ \forall \InHumanVertex \in \InHumanVertices\\
        \sum_{\InHumanVertex \in \InHumanVertices} | \min \big(\SceneSDF(\InHumanVertex), 0 \big) |, & \text{otherwise}
    \end{cases}
\end{equation}
where $\SceneSDF$ is a pre-computed signed distance field (SDF) \cite{Hassan_2019_ICCV} of the scene $\InScene$. When $\SceneSDF(\InHumanVertex)$ has a negative sign, it indicates that the body vertex $\InHumanVertex$ is located inside the nearest scene object (\ie, penetration). Conversely, a positive sign means that $\InHumanVertex$ is positioned on the outside.

\myitem{5)} Finally, to resolve penetration within the human body $\InHuman$ itself, we include a self-penetration energy $\SelfPenetrationEnergy$ based on detecting colliding body triangles using Bounding Volume Hierarchies (BVH) \cite{teschner2005collision}. We refer the reader to \cite{Hassan_2019_ICCV,ballan2012motion,tzionas2016capturing} for more formulation details on $\SelfPenetrationEnergy$.

\subsection{Iterative Refinement}
\label{subsec_iterative_refinement}

A posed 3D human $\InHuman$ interacting with the 3D scene $\InScene$ is generated after applying the VLM-based 2D inpainting (\cref{subsec_multiview_interaction_inpainting}) and the robust 3D lifting (\cref{subsec_3d_interaction_optimization}). To improve the synthesis and consistency of the interaction results, we employ an iterative refinement scheme over the aforementioned inpainting and 3D lifting. During refinement, we render the silhouette of the posed 3D human $\InHuman$ in each camera view, and use it as a more precise and consistent mask $\SDMask$ in the latent diffusion inpainting $\SDInpaint$ \cite{rombach2022high}, replacing dynamic masking. By doing so, the consistency among the 2D pose hypotheses inpainted into the multi-view scene images gradually improves, thus leading to an improved 3D HSI synthesis outcome.

\subsection{Implementation Details}
\label{subsec_implementation}

We implemented \OurName{} with PyTorch \cite{Pytorch:NIPS2019}.
For VLM-based 2D inpainting, we use $\NumSceneImages = 16$ cameras for multi-view rendering. We use 50 denoising steps in Stable Diffusion Inpainting with a state-of-the-art diffusion sampler \cite{lu2022dpm}. 
For dynamic masking during inpainting, we aggregate cross-attention maps with a resolution of $16 \times 16$. 
Input text prompts $\InText$ (\eg, ``sitting on a bench'') are all appended with fixed prefix ``a woman'' and fixed suffix ``wearing a white shirt and blue pants, full body'' to better constrain generation.
For robust 3D lifting, we set $\ViewConsistencyNum = 3$, and optimize the energy $\TotalEnergy$ using gradient descent \cite{loshchilov2018decoupled} for 1.6K steps, which takes $\sim$ 3 minutes on an NVIDIA A100 GPU.
In the iterative refinement, we dilate the rendered human silhouette with a kernel size of $11 \times 11$ for mask generation, and the refinement is performed once.

\section{Experiments}
\label{sec_experiments}

We demonstrate the effectiveness and generality of our approach \OurName{} on a diverse collection of 3D scene models from Sketchfab.com. We conduct both quantitative and qualitative evaluations to compare \OurName{} with alternative  baselines approaches \cite{zhao2022compositional,Hassan_2019_ICCV} to our new task.

\myparagraph{Dataset.} In our Sketchfab dataset, we gathered 8 large-scale 3D scenes encompassing a variety of indoor and outdoor environments with diverse geometric structures, including a realistic Venice city, a gym, and a cartoon-style food truck. We collected 4-5 text prompts per scene describing human interactions with the scene for specified approximate point locations, resulting in 38 actions for evaluation.

\myparagraph{Baselines.}
To the best of our knowledge, there are no baselines that estimate 3D human-scene interactions based on natural language text input in a zero-shot fashion. We thus perform comparisons with related baseline approaches:
\begin{itemize}
    \item COINS \cite{zhao2022compositional} is a state-of-the-art approach estimating 3D humans in indoor 3D scans with a fixed vocabulary of actions and objects, given object segmentations. It takes as input $\braket{\text{action}, \text{object}}$ pairs for semantic control, and requires full supervision in its CVAE training, using captured 3D interaction data with both instance segmentations of 3D objects and action labelling. Due to being trained on a small, closed set of indoor interactions, we adapt COINS to a subset of the most similar Sketchfab actions by manually segmenting corresponding 3D objects from the scenes similar to its indoor training data and mapping the text prompts to its $\braket{\text{action}, \text{object}}$ input. Note that our approach \OurName{} does \emph{not} require any 3D scene segmentations. 
    
    \item Hassan \etal~\cite{Hassan_2019_ICCV} perform 3D human estimation from a single RGB image. To adapt this to Sketchfab, we reuse the multi-view scene images $\{ \SceneHumanImage^{i} \}_{i \in [1, \NumSceneImages]}$ inpainted with humans from our dynamic masking scheme (\cref{subsec_multiview_interaction_inpainting}), where the view with the best image-text cosine similarity (\ie, CLIP score \cite{radford2021clip}) is used as the input for 3D human estimation under the known virtual camera parameters.
    
    \item Ours-Single View: Finally, we consider a baseline leveraging 3D body estimation from our method, but limited to only one inpainted view. The same best view image, as described above, is also as input for this baseline. 
\end{itemize}

\myparagraph{Evaluation Metrics.} To measure 3D HSI quality, we conduct perceptual studies and compute metrics including semantic consistency, diversity, and physical plausibility.

We first carry out two perceptual studies to assess the realism and semantic accuracy of the synthesized interactions. The first is a binary-choice study, where interaction samples generated by two different methods based on the same text prompt are shown, and the participants are asked to choose the sample that is more realistic and better matched the text. The second study is a unary test, where for each interaction sample, the participants are asked to rate the realism and consistency between the shown sample and text prompt from 1 (strongly disagree) to 5 (strongly agree).

To evaluate the semantic consistency between a synthesized 3D interaction and the input text prompt, we calculate the CLIP score \cite{radford2021clip}, where the 3D interaction is re-rendered into $\NumSceneImages$ view images, and the image-text cosine similarities from CLIP ViT-B/32 are averaged across all views.

Additionally, we include quantitative metrics from existing works to evaluate diversity and physical plausibility. 
We observe that these metrics often do not reflect perceptual quality, as generated bodies can be diverse but have interpenetrations without demonstrating any physical or semantic coherence.
Nevertheless, to measure the synthesis diversity, we follow \cite{Zhang_2020_CVPR,zhao2022compositional} and cluster the SMPL-X parameters of the generated humans into 20 clusters with K-means and compute the entropy of the cluster ID histogram of all the samples. We also calculate the cluster size as the mean distance to cluster centers.
For the physical plausibility, we evaluate the collision and contact between body meshes and scene meshes, following  \cite{Zhang_2020_CVPR,zhao2022compositional}. The non-collision score is computed as the ratio between the number of body mesh vertices with positive SDF values (\cref{subsec_3d_interaction_optimization}) and the number of all body mesh vertices. The contact score is defined as the ratio between the number of body meshes with scene contact and the number of all generated body meshes. A body mesh is in contact with the scene if any body mesh vertex has a non-positive SDF value.

\begin{figure}[t]
    \centering
    \includegraphics[width=0.95\columnwidth]{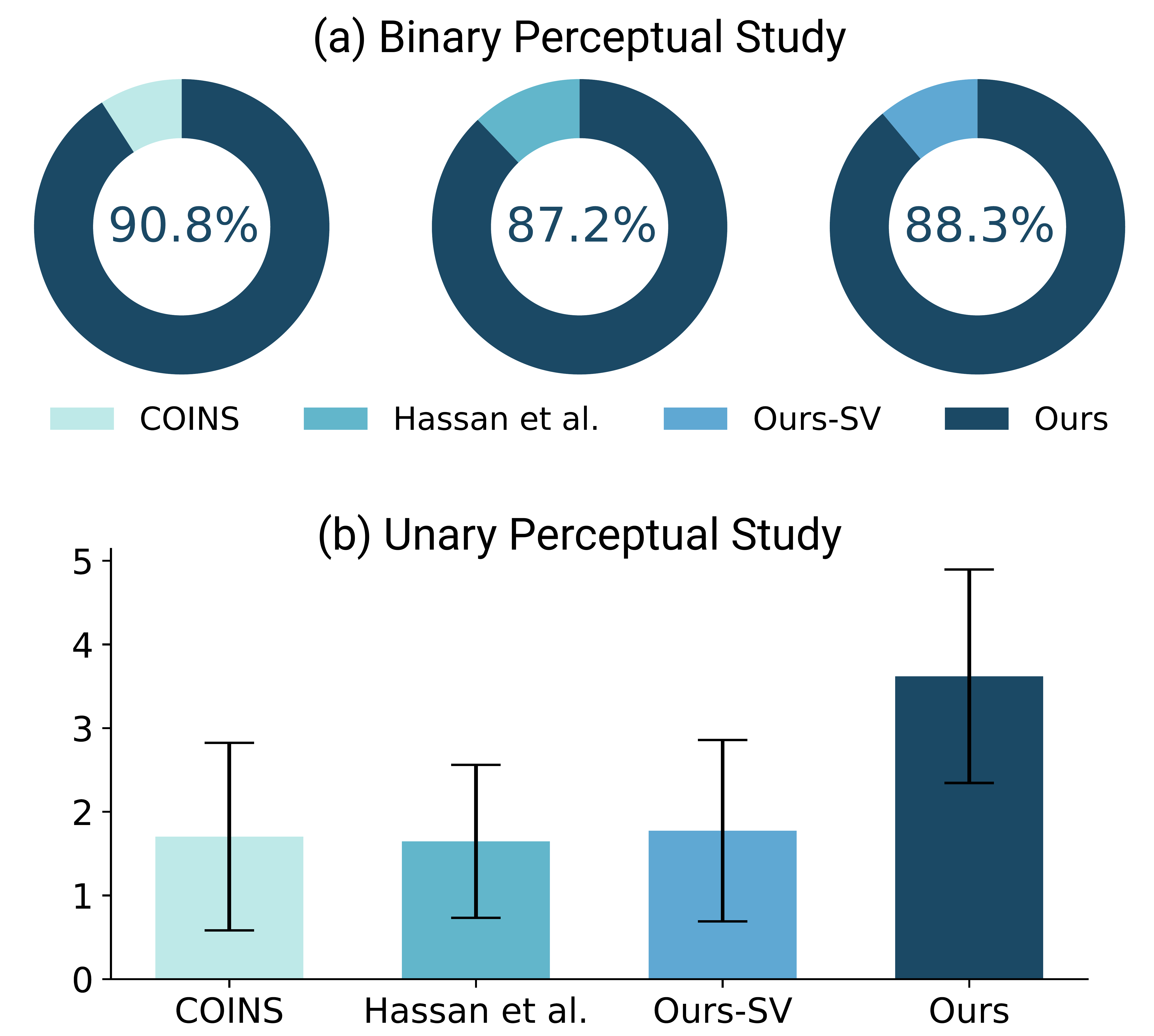}
    \vspace{-0.3cm}
    \caption{User study of 3D human-scene interaction synthesis on the Sketchfab dataset, where participants show a strong preference for the generations by our approach, in comparison with all baselines,  COINS~\cite{zhao2022compositional}, Hassan \etal~\cite{Hassan_2019_ICCV}, and Ours-Single View.}
    \label{fig_userstudy}
\end{figure}

\begin{figure*}[ht!]
    \centering
    \includegraphics[width=0.94\textwidth]{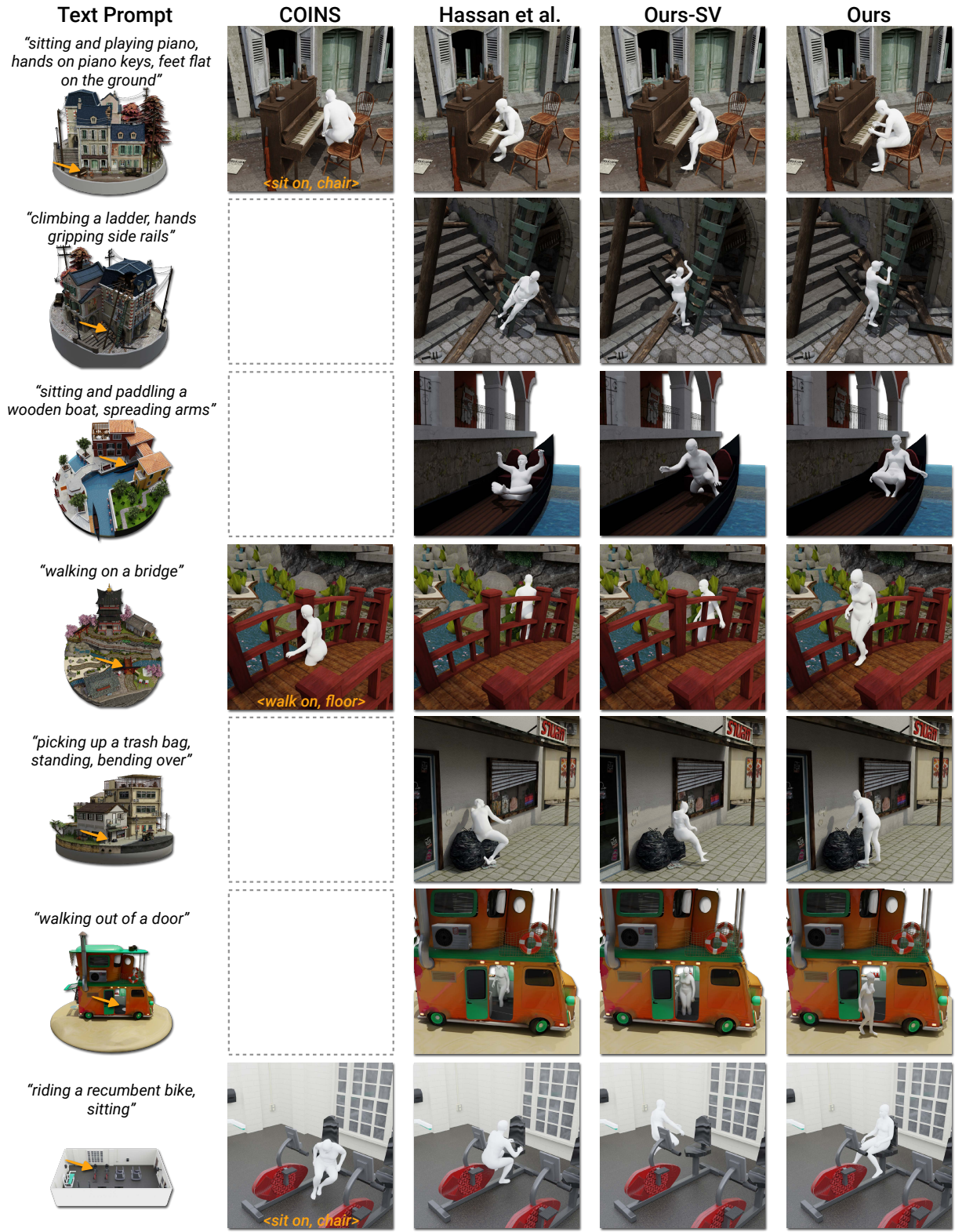}
    \vspace{-0.3cm}
    \caption{Qualitative results on the Sketchfab dataset. Our \OurName{} synthesizes more realistic 3D human-scene interactions and generalizes better across diverse scene types, compared to the baselines COINS~\cite{zhao2022compositional}, Hassan \etal~\cite{Hassan_2019_ICCV}, and Ours-Single View. For COINS, we show the used $\braket{\text{action}, \text{object}}$ labels from its closed set of indoor interactions; its closed setting can lead to degraded results from out-of-distribution classes (e.g., curved floor of bridge, chair at a different height or shape than the train set).}
    \label{fig_qualitative_results_sketchfab}
\end{figure*}

\subsection{Comparison to Alternative Approaches}

\myparagraph{Quantitative Evaluation.}
\cref{fig_userstudy} shows the results of the perceptual studies collected from 30 participants across a binary study and a unary study.
In the binary study, we observe that participants overwhelmingly favor the generations by our \OurName{} compared to all baselines -- more than 87\% of the time.
In the unary test, the average realism rating for our interaction generations is 3.6, the highest compared to the baselines, which are all below 2.0. These perceptual results strongly indicate that our \OurName{} can synthesize realistic 3D humans interacting with various 3D scenes without requiring any captured 3D interaction data.

\cref{tab_sketchfab_comparisons} presents the quantitative evaluation of semantic consistency, diversity, and physical plausibility on the full Sketchfab dataset. Our approach achieves the best semantic consistency score, echoing the strong user preference for \OurName{} in the perceptual studies.
We note that since both Hassan \etal~\cite{Hassan_2019_ICCV} and Ours-Single View operate on single inpainted view samples, results can be very diverse but lack 3D plausibility.
This suggests that the CLIP score is a more reliable metric for assessing the HSI synthesis quality, compared to the diversity and physical plausibility metrics, which reflect less about the generation realism. Nevertheless, our approach has the best diversity entropy and non-collision scores, with the contact score on par.

\begin{table}[ht]
    \centering
    \resizebox{\columnwidth}{!}{%
    \begin{tabular}{lcccccc}
        \hline
                                                        &             Semantics          &                   \multicolumn{2}{c}{Diversity}               &           \multicolumn{2}{c}{Physical Plausibility}           \\
        Method                                          &                CLIP $\uparrow$ &            Entropy $\uparrow$ &       Cluster Size $\uparrow$ &      Non-collision $\uparrow$ &            Contact $\uparrow$ \\
        \hline
        Hassan \etal~\cite{Hassan_2019_ICCV}            &                        0.2598  &                       2.7014  &                       1.1907  &                       0.8824  &                       0.9669  \\
        Ours-SV                                         &                        0.2613  &                       2.6452  &               \textbf{1.5813} &                       0.9765  &               \textbf{1.0000} \\
        Ours                                            &                \textbf{0.2710} &               \textbf{2.7304} &                       1.0500  &               \textbf{0.9767} &                       0.9868  \\
        \hline
    \end{tabular}
    }
    \vspace{-0.2cm}
    \caption{Quantitative comparisons on the Sketchfab dataset. Our approach achieves the best semantic consistency, diversity entropy, and non-collision scores, with the contact score on par.
    Note that single view methods Hassan \etal~\cite{Hassan_2019_ICCV} and Ours-SV tend to produce increased diversity at the cost of semantic plausibility.
    }
    \label{tab_sketchfab_comparisons}
\end{table}

\begin{table}[ht]
    \centering
    \resizebox{0.7\columnwidth}{!}{%
    \begin{tabular}{lccccc}
        \hline
                                    &         w/o DM            &       w/o VC          &       w/o IR                  &           Ours                \\
        \hline
        CLIP $\uparrow$             &         0.2639            &       0.2694          &       0.2664                  &   \textbf{0.2710}             \\
        \hline
    \end{tabular}
    }
    \vspace{-0.2cm}
    \caption{Ablation study on the Sketchfab dataset. The semantic consistency of 3D interaction generations degrades without dynamic masking (DM), view consistency (VC), or iterative refinement (IR), compared to our full approach.}
    \label{tab_sketchfab_ablation}
\end{table}

\begin{figure}[t]
    \centering
    \includegraphics[width=\columnwidth]{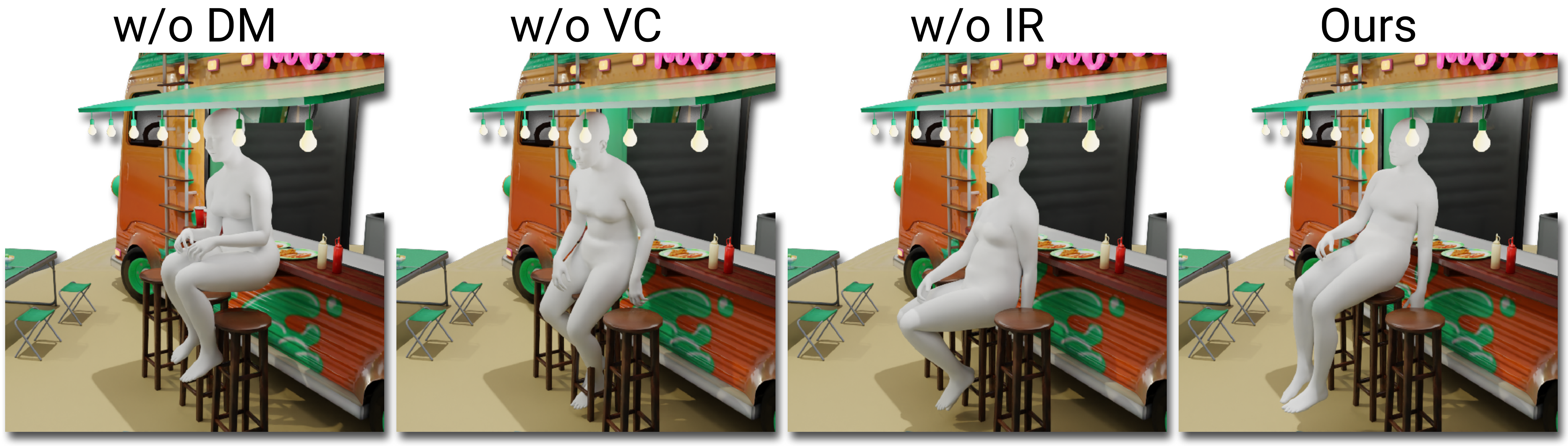}
    \vspace{-0.7cm}
    \caption{Visualization of our method ablations  on Sketchfab dataset for  the input text: ``\emph{sitting on a bar stool}''. 
    Without dynamic masking (DM) or view consistency (VC), the person floats above the middle stool. Without iterative refinement (IR), the person penetrates the stool. Our full approach results in a more realistic synthesis.}
    \label{fig_sketchfab_ablation}
    \vspace{-0.2cm}
\end{figure}

\myparagraph{Qualitative Evaluation.}
We show qualitative comparisons in \cref{fig_qualitative_results_sketchfab}. COINS is severely limited by its training on the closed set of indoor interactions, and thus fails to generalize to outdoor scenes and unseen objects (e.g., no curved floors exist during training, and limited sets of heights and shapes of chairs).
As Hassan \etal~\cite{Hassan_2019_ICCV} and Ours-Single View operate from single views, they both suffer from insufficient pose constraints from other views for plausible interaction generation. In contrast, our approach demonstrates high flexibility and generality to a diverse set of 3D indoor and outdoor scenes by leveraging large VLMs to imagine multi-view interaction hypotheses and then robust 3D lifting.

\subsection{Ablation Studies}
We conduct ablation studies on the Sketchfab dataset to validate the effectiveness of our proposed dynamic masking scheme (\cref{subsec_multiview_interaction_inpainting}), robust 3D lifting with view consistency (\cref{subsec_3d_interaction_optimization}), and iterative refinement (\cref{subsec_iterative_refinement}). Results are presented in \cref{fig_sketchfab_ablation} and \cref{tab_sketchfab_ablation}.

\myparagraph{Dynamic Masking (DM).}
We replace our dynamic masking, used during latent diffusion inpainting, with random masking. We sample a random mask around the image center covering at least 30\% of the image area, and use it as a fixed mask input to the latent diffusion model $\SDInpaint$. We observe that random masking results in noticeably worse quality of interaction synthesis (\cref{tab_sketchfab_ablation}), and incoherence with the scene (\cref{fig_sketchfab_ablation}, floating on the stool). This indicates that our dynamic masking is effective in incorporating sufficient scene context for human inpainting.

\myparagraph{View Consistency (VC).}
We evaluate the role of view consistency in robust 3D lifting by fixing the optimizable scores to $\VarViewsConsistency = \mathbf{1}$. Using all inpainted views leads to averaged, less expressive 3D human poses (\cref{tab_sketchfab_ablation}, \cref{fig_sketchfab_ablation}) due to potential inconsistent 2D pose hypotheses across views. By allowing the solver to adaptively focus on views with consistent inpaintings, our approach generates more realistic 3D HSIs.

\myparagraph{Iterative Refinement (IR).}
Finally, we show the effectiveness of iterative refinement by applying our VLM-based 2D inpainting and robust 3D lifting only once. \cref{tab_sketchfab_ablation} and \cref{fig_sketchfab_ablation} show that iterative refinement improves synthesis quality.

\myparagraph{Limitations.}
Our approach is limited by the inpainting capability of latent diffusion models to imagine possible 2D human-scene compositions, and the diffusion models are also known to be slow at inference time due to their iterative nature \cite{ho2020denoising}. Nevertheless, we believe that our approach can directly benefit from the rapid advancement of VLMs for improved HSI synthesis.

\section{Conclusion}
\label{sec_conclusion}

We have presented the first approach to synthesize general 3D human-scene interactions guided by text inputs. 
Key to our approach is effective distillation of knowledge from large vision-language models, enabling generating 3D humans in scenes without requiring any 3D interaction data for training.
We leverage these powerful vision-language models to generate hypotheses for inpainted 2D human-scene interactions. 
We then formulate a robust optimization to lift the hypotheses to 3D in a view-consistent fashion by simultaneously optimizing for the most informative 2D hypotheses.
Our approach is flexible and can be applied to general scene settings for a variety of actions.
We believe this opens up new opportunities for 3D understanding without requring expensive capture of 3D/4D data.

\myparagraph{Acknowledgements.}
This project is funded by the ERC Starting Grant SpatialSem (101076253), and the German Research Foundation (DFG) Grant ``Learning How to Interact with Scenes through Part-Based Understanding", and supported in part by a Google research gift.

{
    \small
    \bibliographystyle{ieeenat_fullname}
    \bibliography{main}
}

\maketitlesupplementary
\appendix

\setcounter{figure}{6}
\setcounter{table}{2}
\setcounter{equation}{6}

In this supplementary material, we first provide more experimental results in \cref{sec_more_results_supp} and then describe more implementation details in \cref{sec_implementation_supp}.

\section{More Results}
\label{sec_more_results_supp}

\begin{figure*}[bp]
    \centering
    \includegraphics[width=0.94\textwidth]{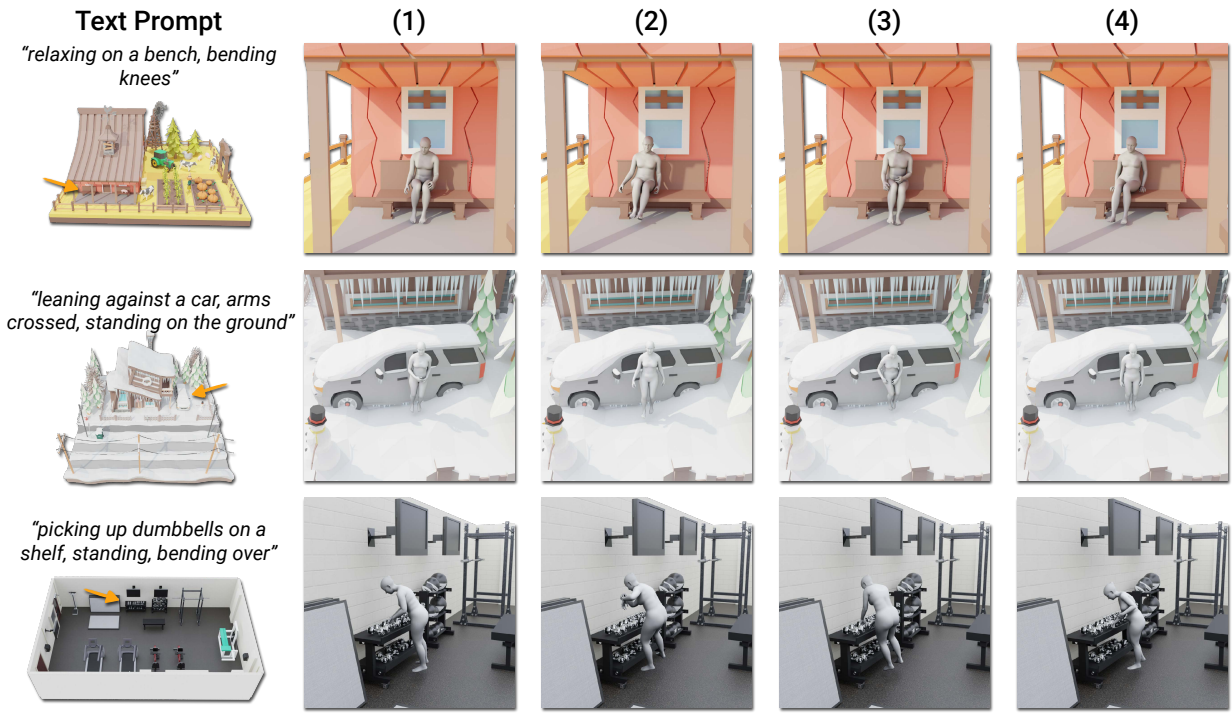}
    \vspace{-0.3cm}
    \caption{Our approach can generate different plausible human interactions in a 3D scene, from (1) to (4), given the same text prompt and location specification.}
    \label{fig_generation_variations}
\end{figure*}

\myparagraph{Generation Variations.}
In \cref{fig_generation_variations}, we show different synthesized 3D interaction variations, given the same 3D scene, text prompt, and location specification input. By using a different collection of multi-view interaction hypotheses, produced by latent diffusion inpainting, our approach can generate various plausible 3D human-scene interactions from the same input.

\begin{figure*}[ht!]
    \centering
    \includegraphics[width=0.94\textwidth]{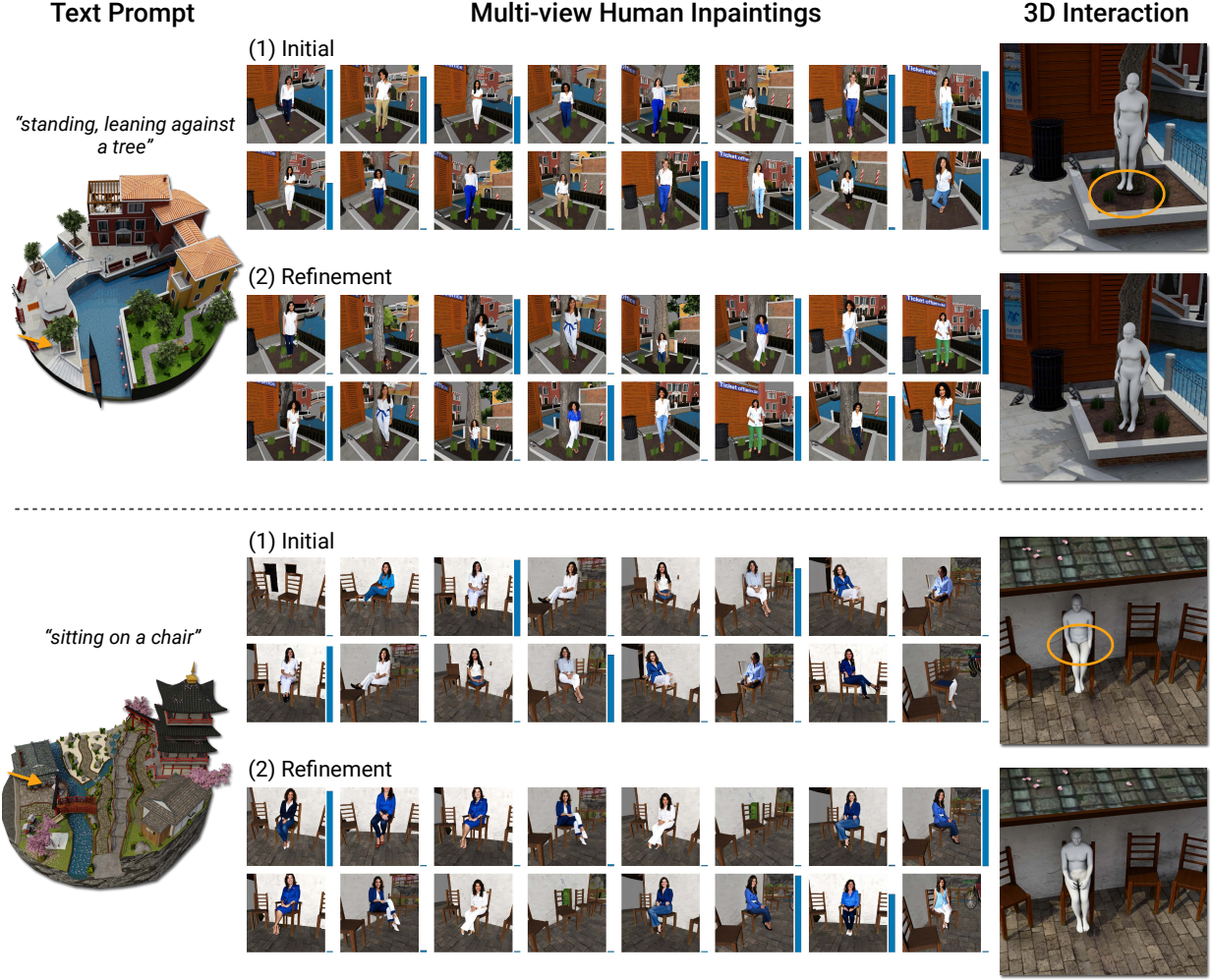}
    \vspace{-0.3cm}
    \caption{Multi-view human inpaintings used in our robust 3D lifting optimization. (1) Initial synthesis stage: images resulting from dynamically-masked inpainting; (2) Refinement  stage: inpaintings with the silhouette masks of the posed 3D human from the initial stage. Without refinement, the person floats above the ground (top), or has self-penetration (bottom). The blue bar next to each inpainted image represents its optimized view consistency score.}
    \label{fig_multiview_inpaintings}
\end{figure*}

\myparagraph{Multi-view human inpaintings.}
In \cref{fig_multiview_inpaintings}, we show the multi-view human inpaintings used in our robust 3D lifting optimization. In the initial synthesis stage, the images are obtained with our dynamically-masked inpainting. In the refinement stage, the images are generated with the silhouette masks of the posed 3D human from the initial stage. The optimized view consistency score is shown next to each inpainted image (as a blue bar). It is observed that through iterative refinement, the quality of both the multi-view human inpaintings and the synthesized 3D interactions gradually improves.

\myparagraph{Evaluation on the PROX-S dataset.}
We perform further comparisons on a recent indoor scene dataset PROX-S \cite{zhao2022compositional} for 3D interaction synthesis. PROX-S consists of 12 indoor scenes with 3D instance segmentations and interaction annotations in the form of $\langle \text{action, object} \rangle$ pairs. Four of the scenes are used for testing. The interaction synthesis is evaluated on about 150 different combinations of action and object instances in the test set.

To adapt our approach \OurName{} to PROX-S, we map the provided interaction labels, \eg, $\langle \text{sit on, sofa} \rangle$, to natural language descriptions, \eg, ``sitting on the sofa''. We use the bounding box centers of object instances as the approximate 3D location input. We stress that in a general synthesis scenario (\eg, scenes from the Sketchfab dataset), our approach does \emph{not} require any 3D scene segmentations.

Several baseline methods focused on indoor 3D interaction synthesis are compared, including PiGraph-X \cite{savva2016pigraphs}, POSA-I \cite{Hassan_2021_CVPR}, and COINS \cite{zhao2022compositional}. Note that all the baselines require learning from the PROX-S training set that has 3D scene segmentations, captured 3D human poses, and interaction labellings. In contrast, our approach does \emph{not} require any 3D learning or captured 3D interaction data.

\begin{figure*}[ht!]
    \centering
    \includegraphics[width=0.94\textwidth]{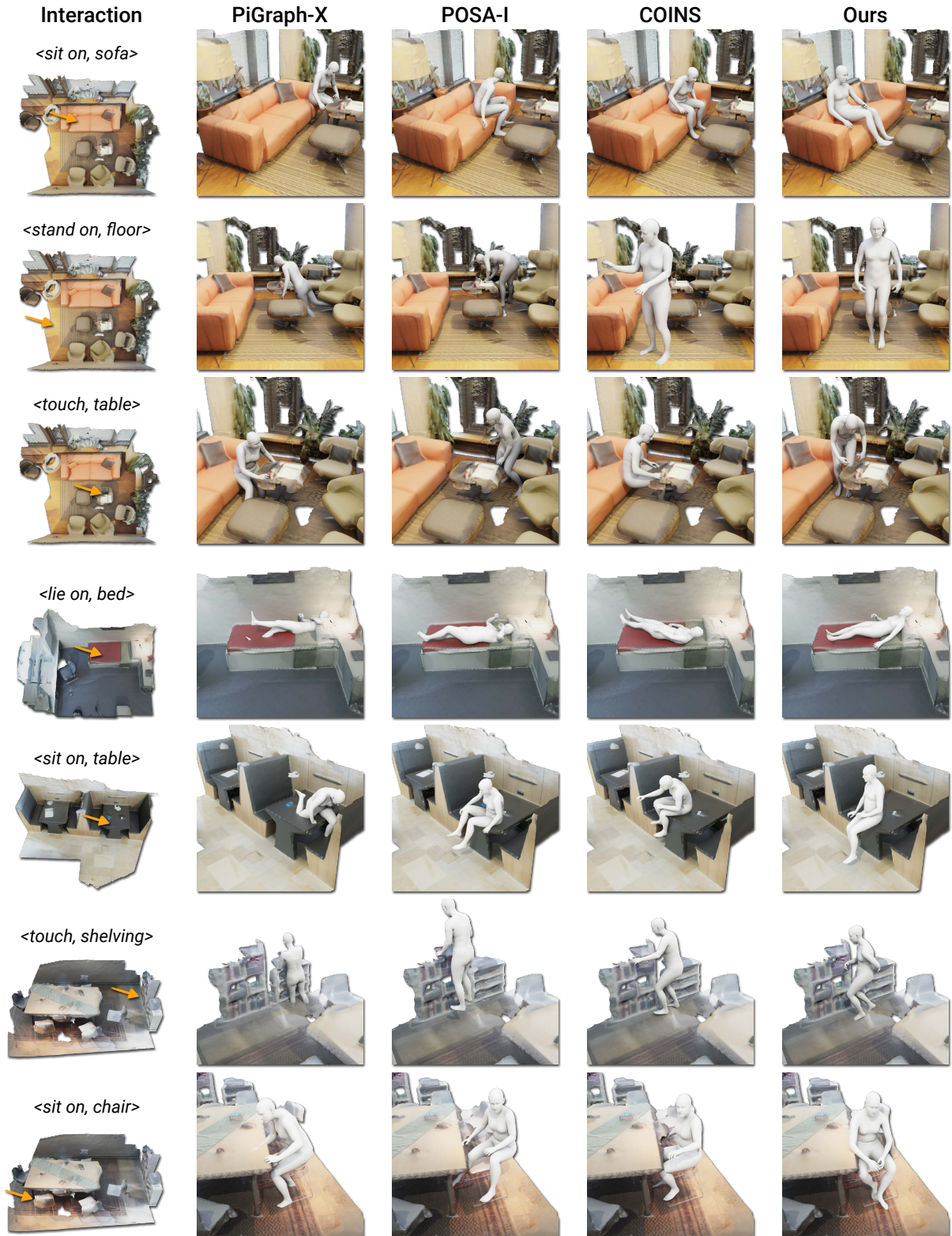}
    \vspace{-0.3cm}
    \caption{Qualitative results on the PROX-S dataset. Our zero-shot approach can synthesize plausible 3D interactions in indoor scenes \emph{without} relying on any captured indoor interaction data. In contrast, the baselines PiGraph-X, POSA-I, and COINS all require \emph{learning} from the PROX-S training set with 3D scene segmentations, captured 3D human poses, and interaction annotations.}
    \label{fig_qualitative_results_proxs}
\end{figure*}

\cref{tab_proxs_comparisons} shows the quantitative evaluation of semantic consistency, diversity, and physical plausibility on PROX-S, and \cref{fig_qualitative_results_proxs} presents the qualitative comparisons. Our zero-shot approach achieves competitive synthesis performance, when compared to the baselines that are specifically trained on PROX-S. We note that the scenes in PROX-S have noisy, incomplete 3D geometry and very low-quality texture details, as shown in \cref{fig_qualitative_results_proxs}, thus their rendered images can be challenging for latent diffusion to inpaint 2D interaction hypotheses. Nevertheless, our approach has the best diversity entropy and non-collision scores, and can generate plausible 3D interactions in indoor scenes without relying on any captured indoor interaction data.

\begin{table}[t]
    \centering
    \resizebox{\columnwidth}{!}{%
    \begin{tabular}{lcccccc}
        \hline
                                                        &                                &             Semantics          &                   \multicolumn{2}{c}{Diversity}               &           \multicolumn{2}{c}{Physical Plausibility}           \\
        Method                                          &                 Zero-Shot      &                CLIP $\uparrow$ &            Entropy $\uparrow$ &       Cluster Size $\uparrow$ &      Non-collision $\uparrow$ &            Contact $\uparrow$ \\
        \hline
        PiGraph-X                                       &                        \xmark  &                        0.2562  &                        3.719  &                        1.019  &                        0.861  &                \textbf{0.981} \\
        POSA-I                                          &                        \xmark  &                        0.2594  &                        3.680  &                \textbf{1.061} &                        0.974  &                        0.941  \\
        COINS                                           &                        \xmark  &                \textbf{0.2617} &                        3.685  &                        0.782  &                        0.981  &                        0.969  \\
        Ours                                            &                        \cmark  &                        0.2544  &                \textbf{3.748} &                        0.869  &                \textbf{0.992} &                        0.961  \\
        \hline
    \end{tabular}
    }
    \caption{Quantitative comparisons on the PROX-S dataset. Our zero-shot approach achieves comparable performance, compared to the baselines that \emph{learn} from the PROX-S training set with 3D scene segmentations, captured 3D human poses, and interaction annotations.}
    \label{tab_proxs_comparisons}
\end{table}

\section{Implementation Details}
\label{sec_implementation_supp}

\myparagraph{Multi-view Camera Setup.}
In Sec.~3.2 of the main text, we create a multi-view representation of the scene context at the location $\InLocation$ by rendering the 3D scene $\InScene$ from $\NumSceneImages$ virtual cameras looking at $\InLocation$. 
To determine the camera positions, we first randomly sample a set of 3D points on the $+z$ hemisphere centered at $\InLocation$ with a radius of $\CamDistance = 2.0$m, assuming $+z$ as the upward direction for the cameras. We then filter these sampled viewpoints according to the visibility of $\InLocation$ via depth testing. For more robust visibility testing, we opt to crop a local surface patch of $\InScene$ at $\InLocation$ within a radius of $\PatchRadius=0.15$m, and compute the ratio of this patch's visible area from each viewpoint, based on which the top-$\NumSceneImages$ viewpoints are selected.

\myparagraph{Dynamic Masking.}
We summarize our dynamic masking scheme in \cref{alg_dynamic_masking} using the same notation as in Sec.~3.2 of the main text. In practice, we use $T=50$ denoising steps in latent diffusion inpainting. We set $T_{\text{min}} = 25$ for updating the mask $\SDMask_{t}$, and keep $\SDMask_{t}$ unchanged after $t < T_{\text{min}}$ to stabilize the inpaintings.

\begin{algorithm}[t]
    \caption{Inpainting with Dynamic Masking}
    \label{alg_dynamic_masking}
    \begin{algorithmic}[1]
        \State \textbf{Input:} An image $\SceneImage$, a text prompt $\InText$, token indices $\SDHumanTokens$
        \State \textbf{Output:} An inpainted image $\SceneHumanImage$
        \State \textbf{Require:} A latent diffusion inpainting model $\SDInpaint$
        \State $\SDNoisyImage_{T} \thicksim \mathcal{N}(0, 1)$ a Gaussian noise latent
        \State $\SDMask_{T} = \mathbf{0}$
        \For{$t = T, T-1, \cdots, 1$}
            \State $\SDNoisyImage_{t-1}, \SDAttentionMaps_{t} \leftarrow \SDInpaint(\SDNoisyImage_{t}, \SDMask_{t}, \SceneImage, \InText, t)$
            \If{$t \geq T_{\text{min}}$}
                \State $\SDMask_{t-1} \leftarrow \text{binarize} \big(\text{sum}(\SDAttentionMaps_{t}[:, h])\big)$
            \EndIf
        \EndFor
        \State \Return $\SDNoisyImage_{0}$
    \end{algorithmic}
\end{algorithm}

\myparagraph{Angle Prior.}
In Sec.~3.3 of the main text, we use an angle prior $\JointAngleEnergy$ to regularize extreme bending of the body joints $\JointAngles \in \mathbb{R}^{21 \times 3}$ represented in the axis-angle form:
\begin{equation}
    \label{eq_joint_angle_energy_supp}
    \JointAngleEnergy = \sum_{j, a \in \Lambda} | \JointAngles_{j, a} | + \sum_{j, a, s \in \Delta} \max(s \cdot \JointAngles_{j, a}, 0),
\end{equation}
where $\JointAngles_{j, a}$ denotes the angle of axis $a$ of the $j$-th joint, and $s$ denotes a sign ($\pm 1$). The 21 body joints are divided into two groups $\Lambda$ and $\Delta$ for different angle regularizations in \cref{eq_joint_angle_energy_supp}, where $\Lambda$ consists of head, feet, and wrists, and the rest joints are included in $\Delta$. More implementation details will be available in our released code upon publication.
 
\myparagraph{Acknowledgements.}
We thank the following artists for generously sharing their 3D scene designs on Sketchfab.com: 
``\href{https://skfb.ly/oGTO6}{a Food Truck Project}'' by xanimvm,
and ``\href{https://skfb.ly/6RoYD}{WW2 Cityscene - Carentan inspired}'' by SilkevdSmissen are licensed under \href{http://creativecommons.org/licenses/by-nc-nd/4.0/}{CC Attribution-NonCommercial-NoDerivs}.
``\href{https://skfb.ly/6GxUO}{Bangkok City Scene}'' by ArneDC,
``\href{https://skfb.ly/6QYJI}{Low Poly Farm V2}'' by EdwiixGG,
``\href{https://skfb.ly/6R6MM}{Low Poly Winter Scene}'' by EdwiixGG,
``\href{https://skfb.ly/6VTUu}{Modular Gym}'' by Kristen Brown,
``\href{https://skfb.ly/6ToLn}{1DAE10 Quintyn Glenn City Scene Kyoto}'' by Glenn.Quintyn, and 
``\href{https://skfb.ly/6TptH}{Venice city scene 1DAE08 Aaron Ongena}'' by AaronOngena are licensed under \href{http://creativecommons.org/licenses/by/4.0/}{Creative Commons Attribution}.

\end{document}